\documentclass[10pt,twocolumn,letterpaper]{article}

\usepackage[pagenumbers]{wacv} 

\usepackage{graphicx}
\usepackage{amsmath}
\usepackage{amssymb}
\usepackage{booktabs}
\usepackage{float}

\usepackage[pagebackref,breaklinks,colorlinks]{hyperref}

\usepackage[capitalize]{cleveref}
\crefname{section}{Sec.}{Secs.}
\Crefname{section}{Section}{Sections}
\Crefname{table}{Table}{Tables}
\crefname{table}{Tab.}{Tabs.}

\def\wacvPaperID{2668} 
\def\confName{WACV}
\def\confYear{2025}

\begin{document}

\title{EfficientCrackNet: A Lightweight Model for Crack Segmentation}

\author{
    Abid Hasan Zim$^{1,*}$ \quad Aquib Iqbal$^{2,*}$ \quad Zaid Al-Huda$^{3}$ \quad Asad Malik$^{4}$ \quad Minoru Kuribayash$^{5}$ \\
    $^{1}$Aligarh Muslim University \quad
    $^{2}$University of Massachusetts Amherst \\
    $^{3}$Stirling College, Chengdu University \quad
    $^{4}$Monash University Malaysia \quad
    $^{5}$Tohoku University \\
}

\maketitle
\begin{abstract}
Crack detection, particularly from pavement images, presents a formidable challenge in the domain of computer vision due to several inherent complexities such as intensity inhomogeneity, intricate topologies, low contrast, and noisy backgrounds. Automated crack detection is crucial for maintaining the structural integrity of essential infrastructures, including buildings, pavements, and bridges. Existing lightweight methods often face challenges including computational inefficiency, complex crack patterns, and difficult backgrounds, leading to inaccurate detection and impracticality for real-world applications. To address these limitations, we propose EfficientCrackNet, a lightweight hybrid model combining Convolutional Neural Networks (CNNs) and transformers for precise crack segmentation. EfficientCrackNet integrates depthwise separable convolutions (DSC) layers and MobileViT block to capture both global and local features. The model employs an Edge Extraction Method (EEM) and for efficient crack edge detection without pretraining, and Ultra-Lightweight Subspace Attention Module (ULSAM) to enhance feature extraction. Extensive experiments on three benchmark datasets Crack500, DeepCrack, and GAPs384 demonstrate that EfficientCrackNet achieves superior performance compared to existing lightweight models, while requiring only 0.26M parameters, and 0.483 FLOPs (G). The proposed model offers an optimal balance between accuracy and computational efficiency, outperforming state-of-the-art lightweight models, and providing a robust and adaptable solution for real-world crack segmentation. 

\renewcommand{\thefootnote}{\fnsymbol{footnote}}
\footnotetext[1]{These authors contributed equally.}

\end{abstract}

\section{Introduction}
\label{sec:intro}

Cracks are common structural faults found on residential buildings, pavements, and bridges. They often result from inadequate load-bearing capacity, compromising safety. Progressive crack propagation impacts structural integrity and longevity. Therefore, identifying and examining cracks is essential for assessing and maintaining structural security ~\cite{chen2023devil, chen2022geometry}. According to the findings of a research that was carried out in 2006, the cost of traffic accidents that were caused by the state of the roads was estimated to be 217.5 billion. This figure represented 43.6\% of the overall cost of accidents that occurred in the United States. ~\cite{zaloshnja2009cost}. The American Society of Civil Engineers (ASCE) classified US road infrastructure in the 2021 Infrastructure Report Card as a "D" grade, indicating poor condition and at risk. An analysis by the U.S. Interstate Highway System reveals that 11\% of interstate highways have pavements that are either in poor or average condition. Specifically, 3\% of the pavements are categorized as poor, while 8\% are categorized as mediocre. Given these critical issues, it is essential to enhance the safety, functionality, and durability of road infrastructure. Effective methods to monitor structural health and assess road conditions enable rapid decision-making and treatment ~\cite{zhou2023deep}.

\begin{figure}[t]
  \centering
  \includegraphics[width= 0.43\textwidth]{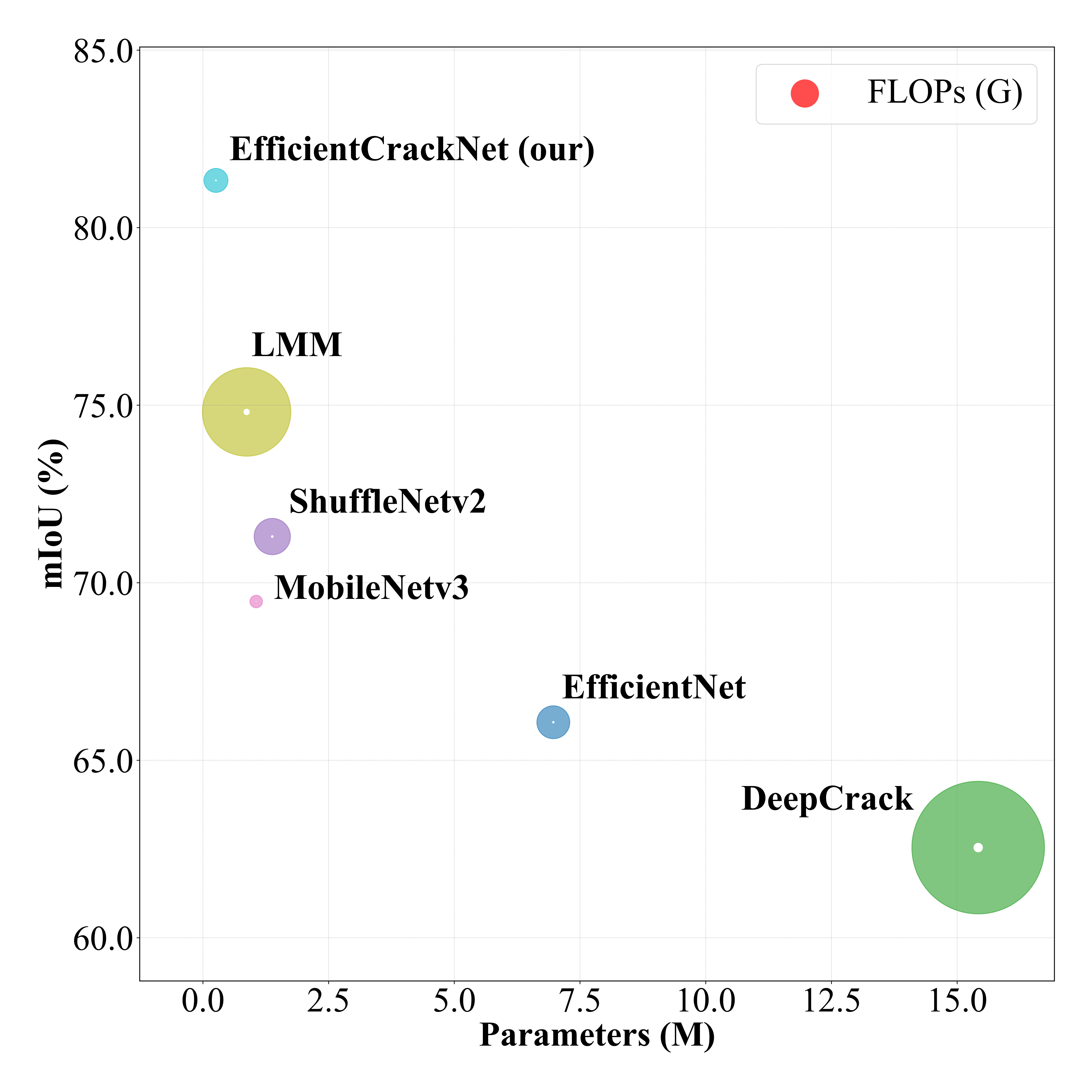}
  \caption{Evaluation of parameters, mIoU, and FLOPs (G) on the Crack500 test dataset. Bubble radius represents the model's size in terms of FLOPs (G).}
  \label{fig:FLOPS}
\end{figure}

Manually crack detection and segmentation is a laborious task that demands highly skilled domain specialists. In order to expedite the advancement of pavement surveys, it is essential to accomplish automated crack segmentation. Crack segmentation is a significant challenge due to uneven intensity, inconsistent contrast, and a very cluttered background. Furthermore, roads of different types have distinct textures that resemble cracks and have very low contrast, therefore complicating the crack recognition process ~\cite{liu2023crackformer, benz2024omnicrack30k, cheng2021joint}. Due to advantages in terms of objectivity, cost-effectiveness, efficiency, and safety, vision-based crack segmentation methods have recently attracted significant interest from both academia and industry ~\cite{zhu2024lightweight, cheng2021joint}. Over the past decade, deep learning has experienced a resurgence, marked by significant successes in various computer vision applications. Most semantic segmentation algorithms currently used in crack segmentation research rely on Convolutional Neural Networks (CNNs) ~\cite{han2021crackw,zhou2023deep, chen2018encoder}. The strength of CNNs lies in their convolutional kernels, which exhibit translation invariance and local sensitivity, allowing them to precisely capture local spatial features. The convolution operation, characterized by its fixed receptive field, is effective at identifying local patterns but is limited in capturing global contextual features or long-range dependencies. In semantic segmentation, relying solely on local features for per-pixel classification can lead to ambiguity, whereas incorporating global contextual features enhances the accuracy of the semantic content for each pixel. However, convolutional kernels have limitations in understanding the overall structure of an image and establishing relationships between different characteristics. Consequently, improving the precision of crack segmentation remains challenging ~\cite{wang2022unetformer}. In contrast, transformer-based networks can capture global contextual features, offering a potential solution to these challenges ~\cite{cao2022swin, cheng2022masked, zim2024tcnformer, iqbal2024eavit, zim2022vision}. Nevertheless, these enhancements in performance are accompanied by sacrifices such as the larger size of models. A significant number of practical applications need the prompt execution of visual recognition tasks on mobile devices with limited resources.

Numerous studies have introduced various deep learning models for crack segmentation, which often depend on high-performance computing devices, such as graphics processing units (GPUs) ~\cite{gong2023state}. These models often have a significant number of parameters and require stable, controlled environments like data centers to ensure reliable operation. However, the practical application of these models in real-world outdoor environments, typical of crack segmentation tasks, is limited due to these dependencies. In contrast, physical-based systems such as drones and mobile robotics have demonstrated significant success in real-world applications. Devices such as the NVIDIA Jetson TX2, Jetson Nano, and Jetson Xavier NX are widely used in edge computing because of their portability, energy efficiency, and compact form factor, making them ideal for real-world applications. These devices can be mounted on unmanned aerial vehicles (UAVs) or robotic platforms to inspect infrastructures such as high-rise buildings, bridges, tunnels, and roadways. Therefore, while deep learning models have achieved precise segmentation performance, their use in practical, on-site applications remains constrained. Edge computing devices offer a more viable solution for these challenging environments ~\cite{zhang2024segment, zhou2022lightweight, wang2023lightweight, li2023automatic}. 

This study introduces a hybrid model combining CNN and transformer architectures for crack segmentation, built upon the UNet framework. The visual comparison between the proposed model and other lightweight segmentation methods on the Crack500 dataset, as shown in Fig. \textcolor{red}{1}, illustrates the relationship between model performance and complexity. Extensive experimental evaluations conducted on three datasets show that EfficientCrackNet strikes an excellent balance between accuracy and computational efficiency. The primary contributions of this research include the following:

\begin{itemize}
    \item We introduced a lightweight hybrid model combining CNN and transformer architectures that effectively balances high accuracy, low computational cost, and efficiency for crack segmentation tasks. The model contains just 0.26 million parameters and 0.483 FLOPs (G).
    \item The Ultra-Lightweight Subspace Attention Module (ULSAM) and MobileViT block are utilized to efficiently capture both local and global features.
    \item An efficient Edge Extraction Method (EEM) is employed for crack edge detection without the need for pretraining.
    \item We performed extensive experiments on three benchmark datasets to validate the effectiveness of the proposed approach.
\end{itemize}

\section{Related Works}
\label{sec:formatting}

Most previous studies on crack segmentation are CNN-based models. U-Net is the most well-known CNN-based model, and various architectures of U-Net have been extensively used for crack segmentation ~\cite{gao2023mra, gao2023pixel, he2024crackham}. Other CNN based models like FCNs ~\cite{wang2020neural, ali2021automatic}, SegNet ~\cite{chen2020pavement, pang2024novel}, DeepLab ~\cite{sun2022dma}, and Mask R-CNN ~\cite{zhang2023automatic} used in crack segmentation in various study. Moreover, various attention modules have also been used with CNN-based models ~\cite{cui2021intelligent}. For example, the attention gate module was integrated into U-Net to enhance the extraction of fracture characteristics by prioritizing important regions and reconstructing semantic features, leading to improved crack segmentation performance ~\cite{konig2019convolutional}. Another study enhanced DeepLabv3+ with a multiscale attention module to combine multiscale crack features ~\cite{sun2022dma}. Similarly, an attention module was incorporated into the DCANet backbone network to combine both detailed and abstract features, enhancing the model's overall performance to recover edge information from the crack ~\cite{qu2022method}. However, CNN-based models often struggle to capture explicit long-range dependencies because of their inherently localized nature. Given that most cracks display narrow, elongated structures, it is essential to capture both local and non-local features for accurate crack segmentation ~\cite{chen2024mind}. Transformer-based models have shown remarkable performance in crack segmentation ~\cite{zhou2024ute, liu2023crackformer}. A study developed SegCrack using MMSegmentation and the OHEM strategy to enhance crack segmentation accuracy ~\cite{wang2022automatic, shrivastava2016training}. Subsequently, another study proposed Crack Transformer ~\cite{guo2023pavement}, a model that integrates elements from the Swin Transformer~\cite{liu2021swin} and SegFormer ~\cite{xie2021segformer}. However, transformer-based models can be initially challenging to train and are prone to overfitting when working with limited datasets ~\cite{liu2023rethinking}. Recently, there has been increasing attention towards hybrid models that combine CNNs and transformers ~\cite{luo2022semi, tao2023convolutional, xu2024sctnet, huang2023semicvt}. Unlike CNNs, transformer has robust long-range modelling capabilities. However, cracks often only occupy a limited part of the picture. Therefore, relying only on a transformer might be susceptible to background interference, resulting in an overall decrease in segmentation performance. Using the hybrid model can compensate for this deficiency ~\cite{xiang2023crack, zhu2023good, xu2023lightweight}. Mobile devices, like drones and phones, often execute crack segmentation activities with limited computational capabilities, memory capacity, and battery support ~\cite{tao2023convolutional}. Hence, our objective is to investigate the use of a compact hybrid model in order to create a network that enhances the precision of crack segmentation with a lightweight structure. 

\section{Methodology}

This section presents a concise overview of the architectural components of the proposed model. The proposed model’s architecture design is based on U-Net. The proposed EfficientCrackNet model consists of three main parts: the encoder, bottleneck, and decoder. The main components of the EfficientCrackNet are the Edge Extraction Method (EEM), Ultra-Lightweight Subspace Attention Mechanism (ULSAM), and MobileViT block.

\subsection{Edge Extraction Method (EEM)}

\begin{figure}[t]
  \centering
  \includegraphics[width= 0.37\textwidth]{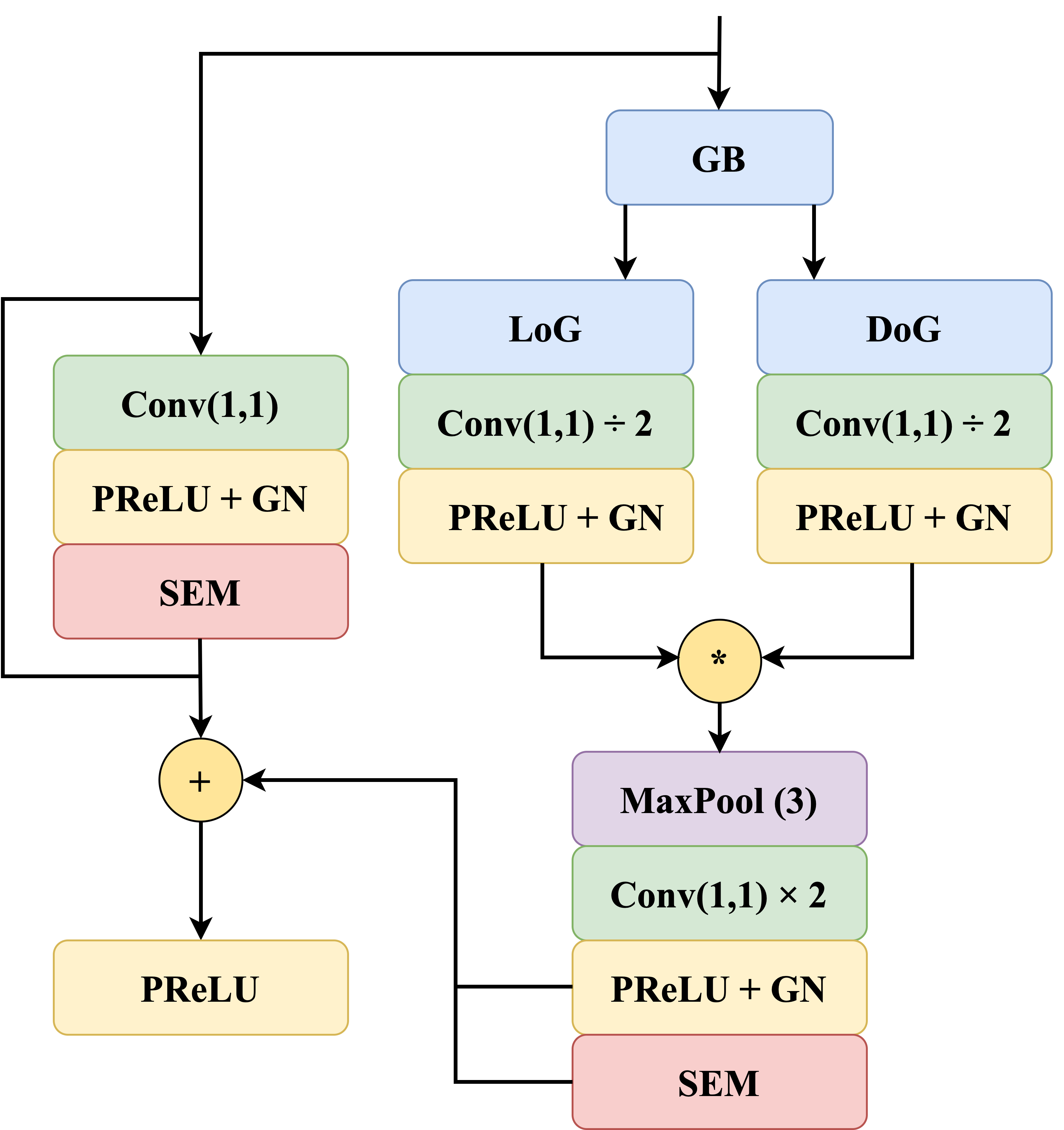}
  \caption{Edge Extraction Method (EEM). }
  \label{fig:onecol}
\end{figure}

\begin{figure*}[t]
  \centering
  \includegraphics[width= .85\textwidth]{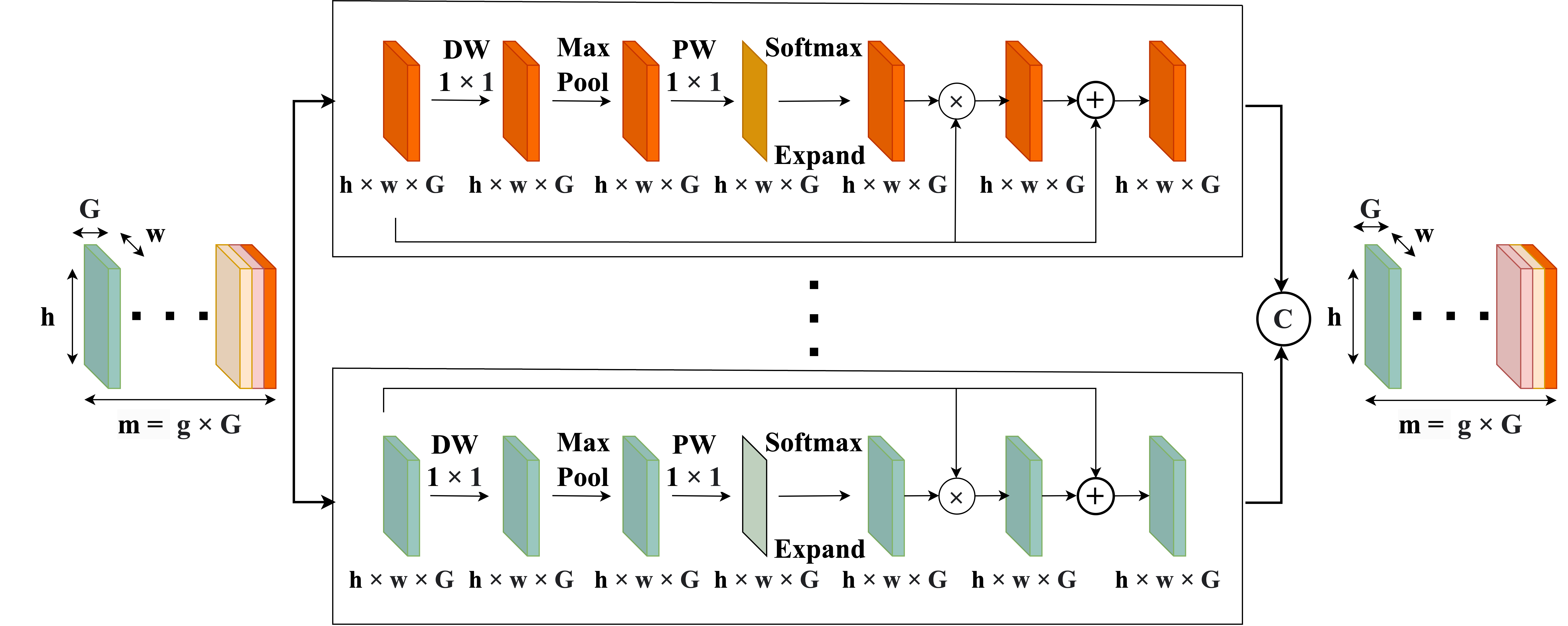}
  \caption{Structure of ULSAM.}
  \label{fig:ULSAM}
\end{figure*}

We employed an Edge Extraction Method (EEM) by combining two conventional edge detection methods, Difference of Gaussian (DoG) and Laplacian of Gaussian (LoG), with convolutional layers. This ultimately resulted in a trainable EEM that was able to learn edge features without the need for separate edge label training and with a minimal number of parameters. For edge extraction, the EEM initially applies Gaussian Blur (GB) to the input by convolving it with a Gaussian kernel of size (3, 3), which smooths out high-frequency components and highlights the overall structure of objects in the image by retaining low-frequency features. Secondly, the mathematical formula of DoG is described by Eq. \textcolor{red}{1}, which involves subtracting the blurred input from the original input to extract the boundaries and borders of objects between the complete image and the low-frequency features in the input. This process is analogous to the band-pass filter effect. Furthermore, combining the Gaussian and Laplacian kernels results in a LoG kernel, as formulated in Eq. \textcolor{red}{2}. When this LoG kernel is convolved with the input image, it performs a second-order derivative, thereby extracting only the edges where there are significant changes. The edges extracted by each method are processed through a (1, 1) convolutional layer independently, then multiplied and passed through a (3, 3) max-pooling layer, followed by another (1, 1) convolutional layer ~\cite{chen2016semantic, al2024lightweight, jing2022recent}. To prevent the loss of edge features, residual connections are incorporated around the EEM, facilitated by the Squeeze-Excitation Module (SEM) ~\cite{hu2018squeeze}. 

\begin{equation}
DoG = I \ast \left(1 - \frac{1}{2 \pi \sigma^2} e^{-\frac{x^2 + y^2}{2 \sigma^2}}\right)
\end{equation}

\begin{equation}
LoG = I \ast \left(-\frac{1}{\pi \sigma^4} \left[1 - \frac{x^2 + y^2}{2 \sigma^2}\right] e^{-\frac{x^2 + y^2}{2 \sigma^2}}\right)
\end{equation}

This integration of SEM within the residual connections accentuates critical features, ensuring the preservation and effective utilization of edge features for crack segmentation. Consequently, incorporating SEM within the residual connections allows the EEM to enhance and maintain edge features, which are vital for precise crack detection and segmentation. The structure of EEM is illustrated in Fig. \textcolor{red}{2}.

\subsection{Ultra-Lightweight Subspace Attention Module (ULSAM)}

The attention mechanism is capable of performing efficient computational modelling of global dependencies and offers an unlimited receptive field. The multi-scale convolution structure has the potential to derive more detailed features; however, interference may result from irrelevant features ~\cite{zheng2017learning, woo2018cbam}. In order to resolve this issue, it is imperative to implement attention mechanisms. The current state-of-the-art attention mechanism is not appropriate for our lightweight model due to its high computational and/or parameter overhead. Therefore, this study employs a simple, effective, and lightweight attention mechanism ULSAM. The ULSAM employs a single attention map for each feature subspace. Initially, ULSAM uses depthwise convolutions (DW) and subsequently applies only one filter during the pointwise convolution (PW) phase to produce the attention maps. This approach significantly reduces computational complexity. 

Let \( F \in \mathbb{R}^{m \times h \times w} \) denote the feature maps from a convolutional layer, where \( m \) is the number of input channels and \( h \) and \( w \) are the spatial dimensions. In ULSAM, the feature maps \( F \) are segmented into \( g \) distinct groups \( [F_1, F_2, \ldots, F_{\tilde{n}}, \ldots, F_g] \), with each group containing \( G \) feature maps. The group \( F_{\tilde{n}} \) represents a specific set of these intermediate feature maps, and the following steps outline the subsequent processing.

\begin{equation}
A_{\tilde{n}} = \text{softmax}(PW^1(\text{maxpool}^{3 \times 3,1} (DW^{1 \times 1}(F_{\tilde{n}}))))
\end{equation}

\begin{equation}
\hat{F}_{\tilde{n}} = (A_{\tilde{n}} \otimes F_{\tilde{n}}) \oplus F_{\tilde{n}}
\end{equation}

\begin{equation}
\hat{F} = \text{concat}([\hat{F}_1, \hat{F}_2, \ldots, \hat{F}_{\tilde{n}}, \ldots, \hat{F}_g])
\end{equation}

In Eq. \textcolor{red}{3}, \( A_{\tilde{n}} \) represents an attention map derived from a group of intermediate feature maps \( F_{\tilde{n}} \). Eq. \textcolor{red}{4} describes the process in which each set of feature maps undergoes refinement to obtain the augmented feature maps \( \hat{F}_{\tilde{n}} \), utilizing feature distribution; here \( \otimes \) denotes element-wise multiplication and \( \oplus \) signifies element-wise addition. The output \( \hat{F} \) produced by ULSAM is obtained by concatenating the feature maps from each groups, as outlined in Eq. \textcolor{red}{5}. This approach allows ULSAM to capture features at multiple scales and frequencies while also facilitating effective cross-channel feature utilization within the network ~\cite{saini2020ulsam}. The structure of ULSAM is illustrated in Fig. \textcolor{red}{3}.

\begin{figure*}[t]
  \centering
  \includegraphics[width= .83\textwidth]{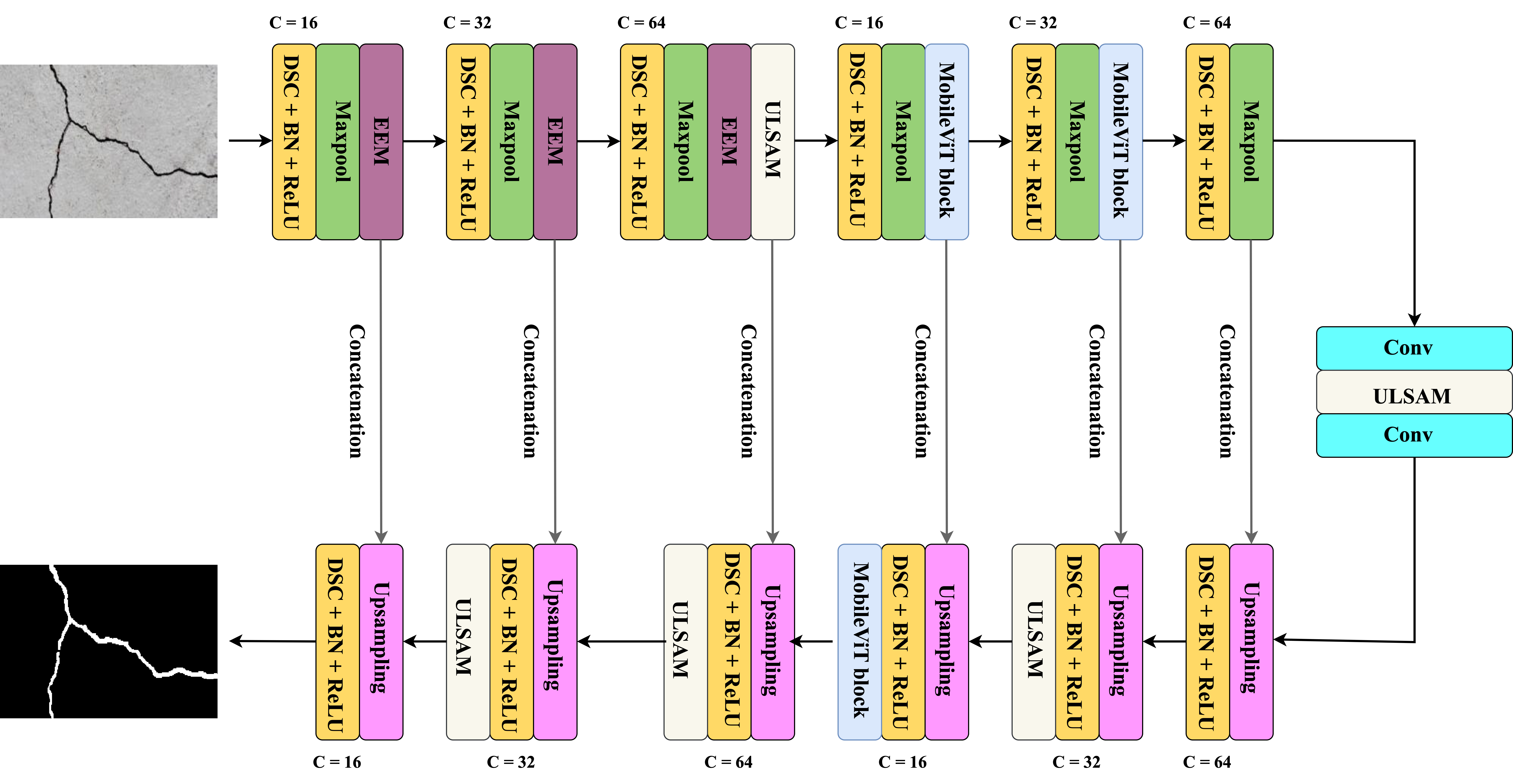}
  \caption{The framework of the EfficientCrackNet.}
  \label{fig:ECN}
\end{figure*}

\subsection{MobileViT block}

The MobileViT block is composed of three distinct sub-modules: local feature encoding, global feature encoding, and feature fusion. Each submodule is tasked with either extracting local features, capturing global features, or merging the extracted features. MobileViT excels at efficiently extracting image feature while maintaining a low parameter count, making it an ideal option for applications constrained by limited computational resources.

Given an input tensor \( \textbf{X} \in \mathbb{R}^{H \times W \times C} \), MobileViT first uses a convolutional layer \( n \times n \), followed by a pointwise (or \( 1 \times 1 \)) convolutional layer, which outputs \( \textbf{X}_L \in \mathbb{R}^{H \times W \times d} \). The \( n \times n \) convolution layer captures local spatial features, while the point-wise convolution maps the tensor to a higher-dimensional space (\( d \)-dimensional, with \( d > C \)) by learning linear combinations of the input channels.

To equip MobileViT with the ability to learn global representations incorporating spatial inductive bias, we divide \( \textbf{X}_L \) into \( N \) non-overlapping flattened patches, denoted as \( \textbf{X}_U \in \mathbb{R}^{P \times N \times d} \). In this context, \( P = wh \), and \( N = \frac{HW}{P} \) represents the number of patches, with \( h \leq n \) and \( w \leq n \) being the height and width of each patch, respectively. For every \( p \in \{1, \cdots, P\} \), inter-patch relationships are captured by utilizing transformers to produce \( \textbf{X}_G \in \mathbb{R}^{P \times N \times d} \) as follows:
\begin{equation}
\textbf{X}_G(p) = \text{Transformer}(\textbf{X}_U(p)), \quad 1 \leq p \leq P
\end{equation}

The \( \textbf{X}_G \in \mathbb{R}^{P \times N \times d} \) is folded to produce \( \textbf{X}_F \in \mathbb{R}^{H \times W \times d} \). Then, \( \textbf{X}_F \) is projected to a lower \( C \)-dimensional space through point-wise convolution and merged with \( \textbf{X} \) through concatenation. A subsequent \( n \times n \) convolutional layer is used to integrate these concatenated features. It is important to note that \( \textbf{X}_U(p) \) captures local features from an \( n \times n \) region using convolutions, while \( \textbf{X}_G(p) \) captures global feature across \( P \) patches for the \( p \)-th location. Consequently, each pixel in \( \textbf{X}_G \) can encompass features from all pixels in \( \textbf{X} \). Therefore, the overall effective receptive field of MobileViT is \( H \times W \).

The MobileViT block improves the network's ability to perceive both global and local features, thereby enhancing its feature extraction capability compared to traditional convolution modules. This convolution-like operation allows the transformer to incorporate positional features, meaning fewer transformer modules are required to learn more features, making it lightweight ~\cite{mehta2021mobilevit}. 

\begin{figure*}[ht!]
  \centering
  \includegraphics[width= .98\textwidth]{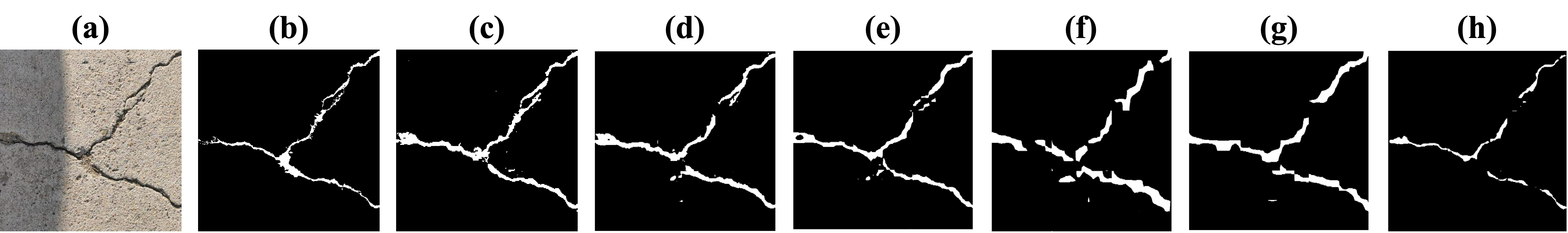}
  \caption{Segmentation results on the DeepCrack dataset (a) Original Image, (b) Ground Truth, (c) EfficientCrackNet (our), (d) LLM, (e) ShuffleNetV2, (f) MobileNetV3, (g) EfficientNet, (h) DeepCrack.}
  \label{fig:RESULT}
\end{figure*}

\subsection{Encoder}

The network is designed to be light, efficient, and robust. To make the model lightweight, in this study, depthwise separable convolutions (DSC) were used. Numerous efficient network architectures utilize DSC as their fundamental components ~\cite{howard2017mobilenets, zhang2018shufflenet, sandler2018mobilenetv2}. DSC notably decreases both the computational load and the total number of parameters in the network, leading to improved efficiency.DW and PW serve distinct purposes in feature generation: DW focuses on identifying spatial relationships, while PW emphasizes capturing correlations across channels ~\cite{guo2019depthwise}. Batch normalization (BN) and ReLU activation are used after each DSC layer. 

The encoder part of the network leverages the EEM, the ULSAM, and the MobileViT block. EEM is incorporated to improve the model's capability in delineating the external boundaries of the cracks. EEM employs a combination of Gaussian and Laplacian filters to extract edges effectively. The module's ability to highlight edges and fine details makes it particularly useful for crack segmentation. ULSAM primarily employs a subspace attention mechanism. This allows the proposed model to capture features at various scales and frequencies while also facilitating effective use of cross-channel features. The convolution operation, characterized by its fixed receptive field, is designed to detect local patterns but inherently struggles to capture global context or long-range dependencies. The MobileViT block addresses this limitation by enabling the model to efficiently encode both local and global features.

\subsection{Decoder}

Like the encoder, the decoder part of the network integrates advanced components such as DSC, upsampling, concatenation blocks, ULSAM, and the MobileViT block to achieve a robust and efficient design. The decoder path begins by increasing the resolution of the feature maps and then combines them with matching feature maps from the encoder, maintaining spatial information. DSCs are also utilized in the decoder to maintain the efficiency and lightweight nature of the network. Each upsampling step is followed by concatenation with the corresponding feature maps from the encoder, allowing the decoder to take advantage of both high-level abstract features and low-level detailed features. This skip connection strategy ensures that the decoder retains important spatial features from the encoder. ULSAM is integrated within the decoder to enhance the learning of cross-channel interdependencies and multiscale features. The MobileViT block is integrated into the decoder to improve the model's capacity for capturing features at both local and global levels.

\section{Experiments and analysis}

This section provides detailed information on the datasets, comparative study, and implementation details.

\begin{table*}
  \centering
  {\small{
  \begin{tabular}{@{}lcccccccccccc@{}}
    \toprule
    Model & \multicolumn{4}{c}{Crack500} & \multicolumn{4}{c}{DeepCrack} & \multicolumn{4}{c}{GAPs384} \\
    \cmidrule(lr){2-5} \cmidrule(lr){6-9} \cmidrule(lr){10-13}
           & Re & Pr & F1 & mIoU & Re & Pr & F1 & mIoU & Re & Pr & F1 & mIoU \\
    \midrule
 
    EfficientNet ~\cite{lo2019efficientdensemodulesasymmetric} & 77.03 & 57.23 & 65.67 & 66.07 & 86.14 & 69.25 & 76.78 & 78.21 & 46.73 & 30.53 & 36.93 & 58.63 \\
    
    DeepCrack ~\cite{liu2019deepcrack} & 44.2 & 73.46 & 55.19 & 62.54 & 61.73 & 94.7 & 74.74 & 77.57 & 3.85 & 65.71 & 7.27 & 50.42\\
    
    ShuffleNetV2 ~\cite{ma2018shufflenetv2practicalguidelines} & 73.64 & 67.70 & 70.54 & 71.30 & 85.01 & 79.71 & 82.27 & 82.72 & 51.22 & 43.63 & 47.12 & 63.26 \\
    
    MobileNetV3 ~\cite{howard2017mobilenets} & 76.28 & 62.49 & 68.7 & 69.47 & 87.01 & 68.40 & 76.59 & 78.04 & 50.43 & 31.09 & 38.46 & 59.11 \\
    LMM ~\cite{al2024lightweight} & \textbf{80.86} & 69.74 & 74.89 & 74.81 & \textbf{ 88.15 } & 84.77 & \textbf{86.43} & 86.43 & 60.66 & 46.65 & 52.74 & 65.87 \\
    
    EfficientCrackNet (our)& 78.43 & \textbf{79.77} &\textbf{79.10} & \textbf{81.33} & 83.37 &\textbf{ 88.54 }& 85.88 & \textbf{87.10} & \textbf{76.87} &\textbf{58.43 }& \textbf{66.40} & \textbf{71.94} \\
    \bottomrule
  \end{tabular}
  }}
  \caption{Comparison results of all models on Crack500, DeepCrack, and GAPs384 test images.}
  \label{tab:results}
\end{table*}

\subsection{Datasets}

The Crack500, DeepCrack, and GAPs384 benchmark datasets is used in this study.

\textbf{Crack500:} The Crack500 dataset consists of 447 images at a resolution of 2560 × 2592 pixels, featuring diverse crack shapes and widths against complex background textures, making segmentation challenging ~\cite{zhang2016road}.

\textbf{DeepCrack:} DeepCrack is a widely recognized dataset used for assessing crack detection algorithms. It consists of 537 images, each with a resolution of 384 × 544 pixels. The dataset features clear differences in intensity between the cracks and the background, which aids in the effective identification of cracks in pavement images ~\cite{liu2019deepcrack}.

\textbf{GAPs384:} The GAPs384 dataset features 384 high-resolution images (1080 × 1920 pixels) with diverse noise types and complex road textures, making crack segmentation challenging and crucial for developing advanced algorithms ~\cite{yang2019feature}.

\subsection{Data Augmentation}

To enhance the generalization of our model, we utilized various data enhancement techniques. The images were augmented using flipping (\(p = 0.7\)), rotation (\(p = 0.7\)), random brightness and contrast adjustments (\(p = 0.2\)), Gaussian blurring (\(p = 0.2\)) and shift scale-rotate transformations (\(p = 0.2\)). Additionally, Gaussian noise (\(p = 0.2\)), color inversion (\(p = 0.2\)) were applied. These augmentations simulate various real-world conditions, enhancing the model's robustness and ability to generalize effectively across different scenarios.

\subsection{Loss Function}

The Dice Coefficient Loss is used in this study. It is defined as:

\begin{equation}
\text{Dice Coefficient} = \frac{2|A \cap B|}{|A| + |B|}
\end{equation}

Where \(A\) is the set of predicted pixels and \(B\) is the set of ground truth pixels. To use the Dice Coefficient as a loss function, we define the Dice Loss as:

\begin{equation}
\text{Dice Loss} = 1 - \text{Dice Coefficient}
\end{equation}

In practice, this can be written for continuous variables as:

\begin{equation}
\text{Dice Loss} = 1 - \frac{2 \sum_{i=1}^{N} p_i g_i}{\sum_{i=1}^{N} p_i^2 + \sum_{i=1}^{N} g_i^2}
\end{equation}

where \(p_i\) and \(g_i\) represent the predicted and ground truth values for pixel \(i\), respectively, and \(N\) is the total number of pixels \cite{9277638, zhao2020rethinking}.

\subsection{Evaluation metrics}

Four widely recognized evaluation metrics are employed: Recall (Re), Precision (Pr), F-score (F1), and the mean Intersection-over-Union (mIoU). These metrics are defined as follows: 

\begin{equation}
Re = \frac{TP}{TP + FN}
\end{equation}

\begin{equation}
Pr = \frac{TP}{TP + FP}
\end{equation}

\begin{equation}
F1 = 2 \times \frac{Re \times Pr}{Re + Pr}
\end{equation}

\begin{equation}
mIoU = \frac{1}{n_{cl}} \frac{\sum_i n_{ii}}{t_i + \sum_j n_{ji} - n_{ii}}
\end{equation}

The quantities TP, FP, FN, and TN represent the following: true positives, false positives, false negatives, and true negatives, respectively. Here, \( n_{ij} \) denotes the number of pixels classified as class \( j \) in class \( i \), \( n_{cl} \) represents the number of classes, and \( t_i \) is the total number of pixels in class \( i \), calculated as \( t_i = \sum_{j} n_{ij} \).

\subsection{Comparison with the lightweight models}
We evaluate the performance of several state-of-the-art and lightweight segmentation models with our model on three datasets. The models compared are 
EfficientNet \cite{lo2019efficientdensemodulesasymmetric}, DeepCrack ~\cite{liu2019deepcrack}, ShuffleNetV2 \cite{ma2018shufflenetv2practicalguidelines}, MobileNetV3 \cite{howard2017mobilenets}, LMM ~\cite{al2024lightweight}. Here, Table \textcolor{red}{1} shows the results. Fig. \textcolor{red}{5} compares the segmentation outputs of our model and other lightweight models to the ground truth on the DeepCrack dataset, demonstrating the robustness of the proposed model. 

\textbf{The Results on the Crack500 dataset}: On the Crack500 test dataset, the proposed EfficientCrackNet demonstrated a mIoU of 81.33\%, Re of 78.43\%, and Pr of 79.77\%. In terms of the F1-score, which evaluates the balance between Pr and Re, EfficientCrackNet achieved the highest score of 79.10\%. In comparison to other models, EfficientCrackNet demonstrated significantly superior performance, achieving a mean Intersection over Union (mIoU) that was 8.02\% higher than LMM, 17.07\% higher than MobileNetV3, 14.07\% higher than ShuffleNetV2, 18.79\% higher than DeepCrack, and 30.04\% higher than EfficientNet.

\textbf{The Results on the DeepCrack dataset:} When tested on the DeepCrack dataset, the proposed EfficientCrackNet model achieved a mIoU of 87.10\%, a Re of 83.37\%, and a Pr of 88.54\%. In terms of the F1-score, EfficientCrackNet reached 85.88\%. Comparative analysis reveals that EfficientCrackNet substantially outperforms other models, with a notable mIoU improvement of 0.77\% over LMM, 11.61\% over MobileNetV3, 5.29\% over ShuffleNetV2, 12.29\% over DeepCrack, and 11.37\% over EfficientNet.

\textbf{The Results on the GAPs384 dataset:} On the GAPs384 test dataset, EfficientCrackNet achieved a Re of 76.87\%, a Pr of 58.43\%, and an F1-score of 66.40\%. Additionally, it obtained a mIoU of 71.94\%, which represents the highest value among the models evaluated. In comparison to other models, EfficientCrackNet demonstrated superior performance, with an mIoU that was 8.44\% higher than LMM, 21.71\% higher than MobileNetV3, 13.72\% higher than ShuffleNetV2, 29.91\% higher than DeepCrack, and 22.70\% higher than EfficientNet.

The consistently better performance of the proposed EfficientCrackNet indicates its robustness and adaptability while maintaining computational efficiency.

\begin{table}[h!]
  \centering
  {\small{
  \begin{tabular}{@{}lcccccc@{}}
    \toprule
    Model & FLOPs (G) & Parameters (M) \\
    \midrule
    EfficientNet \cite{lo2019efficientdensemodulesasymmetric} & 0.906 & 6.97 \\
    DeepCrack ~\cite{liu2019deepcrack} & 15.42 & 14.72 \\
    ShuffleNetV2 \cite{ma2018shufflenetv2practicalguidelines} & 1.105 & 1.38 \\
    MobileNetv3 \cite{howard2017mobilenets} & \textbf{0.132} & 1.06 \\
    LMM ~\cite{al2024lightweight} & 6.56 & 0.87 \\
    EfficientCrackNet (our) & 0.483 & \textbf{0.26} \\
    \bottomrule
  \end{tabular}
  }}
  \caption{Comparison of models complexity in terms of FLOPs (G) and Parameters (M).}
  \label{tab:model_complexity}
\end{table}

\subsection{Model Complexity}

Lightweight networks seek to reduce the complexity of models by addressing three key factors: floating-point operations per second (FLOPs), and the number of parameters. The FLOPs and parameters can be quantified using the formulas outlined in Eq. \textcolor{red}{14} and \textcolor{red}{15}.

\begin{equation}
FLOPs = 2HW(C_{in} K^2 + 1)C_{out}
\end{equation}
\begin{equation}
Params = (C_{in} K^2 + 1)C_{out}
\end{equation}

In this context, \( H \) and \( W \) refer to the height and width of the input feature map, respectively. \( C_{in} \) represents the number of input channels, and \( C_{out} \) signifies the number of output channels. The variable \( K \) corresponds to the kernel size.

Table \textcolor{red}{2} presents the FLOPs (G) and parameters (M) of the models. Our model, EfficientCrackNet, has the fewest parameters, with 0.61M and 0.80M fewer than LMM and MobileNetv3, respectively, due to its efficient design. EfficientCrackNet has the second-lowest FLOPs (G), which is 0.483, with only MobileNetv3 having fewer FLOPs (G). However, EfficientCrackNet significantly outperforms MobileNetv3 across all three datasets. Despite being lightweight, our model performs well on all three datasets, indicating its suitability for real-world crack segmentation on mobile devices.

\begin{figure}[t]
  \centering
  \includegraphics[width=0.48\textwidth]{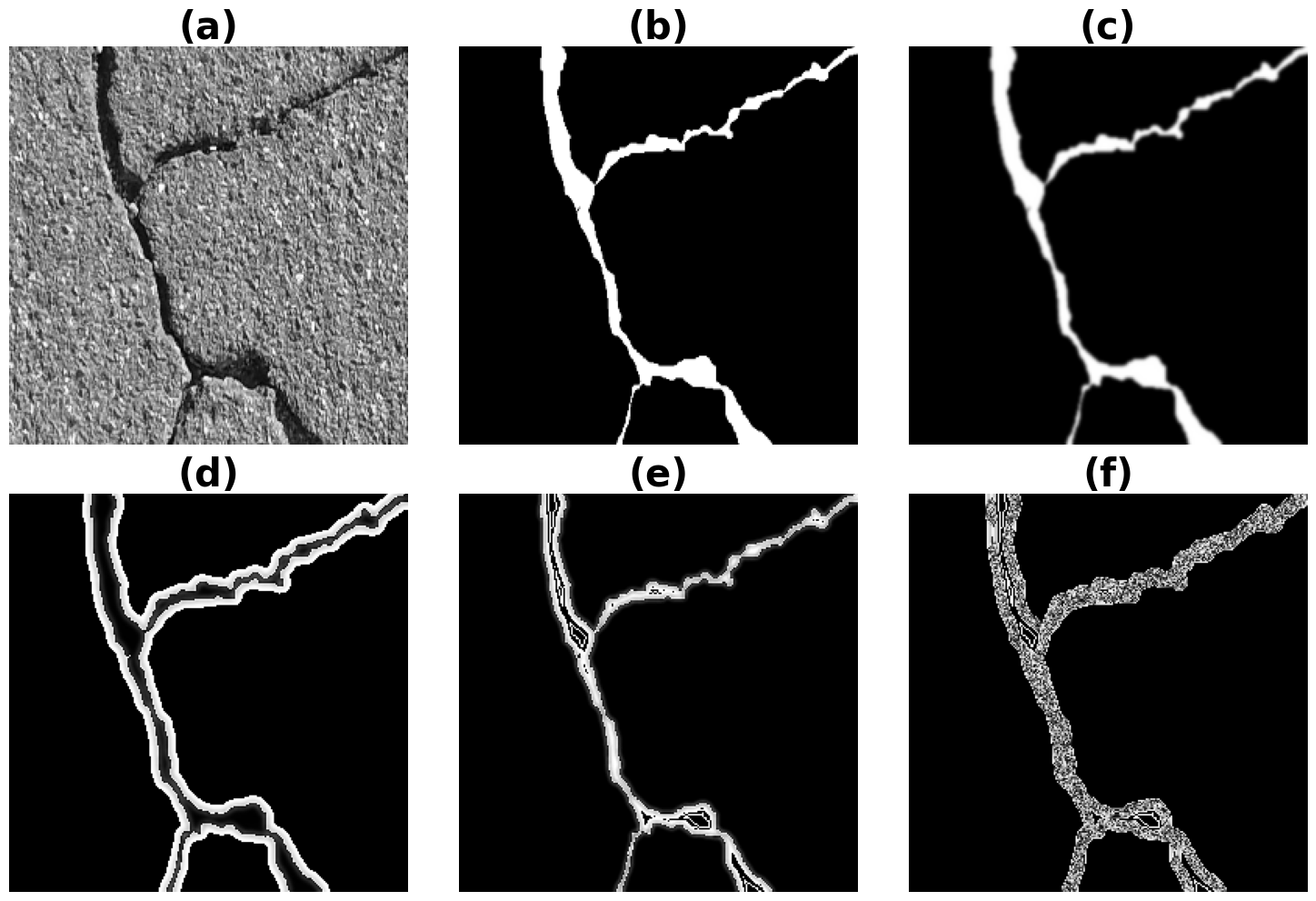}
  \caption{ Feature maps of each part of EEM, (a) Original Image, (b) Segmentation Result, (c) Gaussian Blur (GB), (d) Different of Gaussian (DoG), (e) Laplacian of Gaussian (LoG), (f) Multiplied DoG and LoG.}
  \label{fig:EEM}
\end{figure}

\section{Ablation Study}

This section discusses the effect of ULSAM, MobileViT block, and EEM on our model. The Crack500 dataset is used as an illustrative case due to its wide range of shapes and widths, as well as the intricate background textures present in most of the images.

\subsection{Effect of ULSAM and MobileVit block}

We conducted experiments to evaluate the impact of ULSAM and the MobileViT block within our model. The results, presented in Table \textcolor{red}{3}, are based on the Crack500 dataset. Without ULSAM resulted in a reduction of the F1 score by 1.49\% and the mIoU by 1.18\%. Similarly, the exclusion of the MobileViT block led to a decrease in the F1 score by 4.79\% and the mIoU by 3.40\%. ULSAM is integrated within the encoder, bottleneck, and decoder of the model. ULSAM improve the network's ability to understand complex visual patterns, without significantly increasing model parameters. On the other hand, the MobileViT block aids in capturing global features without escalating the model's complexity significantly.

\begin{table}[h!]
\centering
\begin{tabular}{@{}lcccccc@{}}
\toprule
Baseline & F1 & mIoU \\ 
\hline

       w/o ULSAM  & 77.93& 80.38 & \\ 
   w/o MobileVit block & 75.40 & 78.61 & \\ 
      \textbf{Our} & \textbf{79.10} & \textbf{81.33} & \\ 
 
\bottomrule
\end{tabular}
\caption{Impact of ULSAM and MobileVit block.}
\end{table}

\subsection{Effect of EEM}

\begin{table}[h!]
\centering
\begin{tabular}{@{}lcccccc@{}}
\toprule
Baseline & F1 & mIoU \\ 
\hline

       w/o EEM  & 71.06 & 75.47 & \\ 
      \textbf{Our} & \textbf{79.10} & \textbf{81.33} & \\ 
 
\bottomrule
\end{tabular}
\caption{ Influence of EEM on Crack500 datasets.}
\end{table}

Edges and object boundaries are crucial for various high-level computer vision tasks, such as image segmentation. In the early layers of a model, feature maps retain more spatial details of the original shapes of objects, making the extraction of these edges particularly important for tasks like crack segmentation. In our proposed model, we employ an EEM in the earlier layers of the encoder. We conducted experiments to evaluate the impact of EEM on our model using the Crack500 dataset. The results, presented in Table \textcolor{red}{4}, show that the exclusion of EEM led to a significant reduction in performance, with the F1 score dropping by 10.71\% and the mIoU by 7.48\%. This highlights the importance of EEM in enhancing segmentation accuracy. Fig. \textcolor{red}{6} illustrates the effect of each component of EEM on the segmentation mask.

\section{Conclusion}

EfficientCrackNet is a lightweight hybrid model designed for automatic crack detection and segmentation in infrastructure maintenance. It combines DSC and MobileViT blocks to capture global and local features, enhancing segmentation precision. The model uses an innovative EEM with DoG and LoG for feature extraction without additional training and incorporates ULSAM for improved feature representation. EfficientCrackNet achieves state-of-the-art results on three benchmark datasets, outperforming other lightweight models with just 0.26M parameters and 0.483 GFLOPs, making it ideal for real-world applications.

Despite its strengths, EfficientCrackNet has limitations that require further exploration. It struggles with detecting extremely thin cracks, which may need more advanced feature extraction techniques. Additionally, variations in lighting and background conditions can affect its performance. Future research should refine the model and expand its use in structural health monitoring, enhancing automated inspection for better infrastructure maintenance and safety.

{\small
\bibliographystyle{ieee_fullname}
\bibliography{wacv}
}

\end{document}